\documentclass[10pt,twocolumn,letterpaper]{article}
\usepackage{times}
\usepackage{epsfig}
\usepackage{graphicx}
\usepackage{amsmath}
\usepackage{amssymb}
\usepackage{booktabs}
\usepackage{epigraph}

\usepackage{bm}
\usepackage{multirow}
\makeatletter

\newcommand{\Rmnum}[1]{\expandafter\@slowromancap\romannumeral #1@}
\makeatother
\usepackage[table]{xcolor}
\usepackage{utfsym}
\usepackage[accsupp]{axessibility}

\usepackage[pagenumbers]{cvpr} 

\definecolor{cvprblue}{rgb}{0.21,0.49,0.74}
\usepackage[pagebackref,breaklinks,colorlinks,citecolor=cvprblue]{hyperref}

\def\confName{CVPR}
\def\confYear{2024}

\title{Robust Emotion Recognition in Context Debiasing}

\author{Dingkang Yang$^{1,2}$ $\quad$
        Kun Yang$^{1}$ $\quad$
        Mingcheng Li$^{1,2}\quad$
        Shunli Wang$^{1,2}\quad$ \\
        Shuaibing Wang$^{1,2}\quad$ 
        Lihua Zhang$^{1,2,3,4}$\footnotemark[4] \\ 
        \small$^1$Academy for Engineering and Technology, Fudan University$\,$ 
        \small$^2$Cognition and Intelligent Technology Laboratory (CIT Lab)\\
        \small$^3$Jilin Provincial Key Laboratory of
Intelligence Science and Engineering, Changchun, China\\
\small$^4$Engineering Research Center of AI and Robotics, Ministry of Education, Shanghai, China\\
{\tt\small \{dkyang20,\,lihuazhang\}@fudan.edu.cn}
}

\begin{document}
\maketitle

\renewcommand{\thefootnote}{\fnsymbol{footnote}}
\footnotetext[4]{Corresponding author.}

\begin{abstract}
Context-aware emotion recognition (CAER) has recently boosted the practical applications of affective computing techniques in unconstrained environments. Mainstream CAER methods invariably extract ensemble representations from diverse contexts and subject-centred characteristics to perceive the target person's emotional state. 
Despite advancements,
the biggest challenge remains due to context bias interference. The harmful bias forces the models to rely on spurious correlations between background contexts and emotion labels in likelihood estimation, causing severe performance bottlenecks and confounding valuable context priors. In this paper, we propose a \underline{c}ounterfactua\underline{l} \underline{e}motion in\underline{f}erence (CLEF) framework to address the above issue. Specifically, we first formulate a generalized causal graph to decouple the causal relationships among the variables in CAER. Following the causal graph, CLEF introduces a non-invasive context branch to capture the adverse direct effect caused by the context bias. During the inference, we eliminate the direct context effect from the total causal effect by comparing factual and counterfactual outcomes, resulting in bias mitigation and robust prediction.
As a model-agnostic framework, CLEF can be readily integrated into existing methods, bringing consistent performance gains.
\end{abstract}

\section{Introduction}
\label{sec:intro}
\setlength{\epigraphwidth}{0.45\textwidth} 
\epigraph{\emph{``Context is the key to understanding, but it can also be the key to misunderstanding.''}}{\footnotesize--\emph{Jonathan Lockwood Huie}}
\vspace{-3pt}

\begin{figure}[t]
  \centering
  \includegraphics[width=\linewidth]{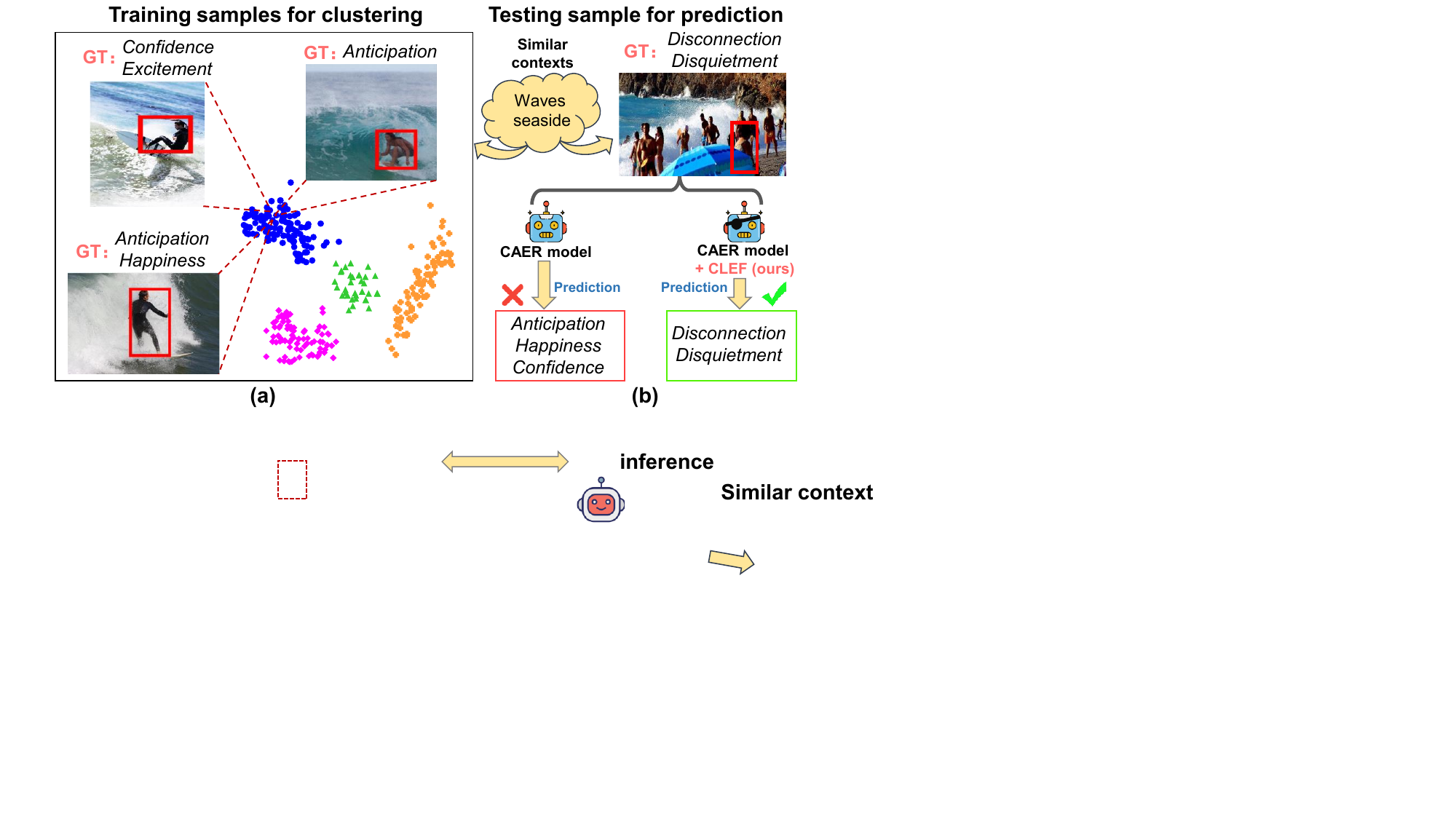}
  \caption{Illustration of the context bias in the CAER task. GT stands for the Ground Truth. Context-specific semantics easily yield spurious shortcuts with emotion labels during training to confound the model~\cite{mittal2020emoticon}, giving erroneous results. Conversely, our CLEF effectively corrects biased predictions.
  }
  \label{intro1}
\vspace{-6pt}
\end{figure}

As the spiritual grammar of human life, emotions play an essential role in social communication and intelligent automation~\cite{lei2023text}.
Accurately recognizing subjects' emotional states from resource-efficient visual content has been extensively explored in various fields, including online education~\cite{jacobs2014emotion}, driving monitoring~\cite{yang2023aide}, and human-computer interaction~\cite{alnuaim2022human}. Conventional works have focused on extracting emotion-related information from subject attributes, such as facial expressions~\cite{du2021learning}, body postures~\cite{bhattacharya2020step}, acoustic behaviors~\cite{lian2021ctnet}, or multimodal combinations~\cite{mittal2020m3er,yang2023target,yang2022learning,yang2022contextual,yang2022disentangled,li2023towards}. Despite considerable advances in subject-oriented efforts, their performance suffers from severe bottlenecks in uncontrolled environments.
As shown in \Cref{intro1}\textcolor{red}{a}, physical representations of subjects in wild-collected images are usually indistinguishable 
(\eg, ambiguous faces) due to natural occlusions that fail to provide usable emotional signals.

Inspired by psychological research~\cite{barrett2011context},  context-aware emotion recognition (CAER)~\cite{kosti2017emotion} has been proposed to seek additional affective semantics from situational contexts. The contexts~\cite{kosti2019context} are typically considered to include out-of-subject factors, such as background objects, place attributes, scene elements, and dynamic interactions of surrounding agents. These rich contextual stimuli promisingly provide complementary emotion clues for accurate recognition.
Most existing methods perform emotion inference by extracting ensemble representations from subjects and contexts using sophisticated structures~\cite{mittal2020emoticon,mittal2021multimodal,zhang2019context,li2021human,yang2022emotion,wang2022context}or customized mechanisms~\cite{ruan2020context,li2021sequential,hoang2021context,chen2023incorporating,de2023high,gao2021graph}. Nevertheless, a recent study~\cite{yang2023context} found that CAER models tend to rely on spurious correlations caused by a context bias rather than beneficial ensemble representations. An intuitive illustration is displayed in \Cref{intro1}. We first randomly choose some training samples on the EMOTIC dataset~\cite{kosti2019context} and perform unsupervised clustering. From \Cref{intro1}\textcolor{red}{a}, samples containing seaside-related contexts form compact feature clusters, confirming the semantic similarity in the feature space.
These samples have positive emotion categories, while negative emotions are nonexistent in similar contexts. In this case, the model~\cite{mittal2020emoticon} is easily misled to capture spurious dependencies between context-specific semantics and emotion labels. In the testing phase from \Cref{intro1}\textcolor{red}{b}, oriented to the sample with similar contexts but negative emotion categories, the model is confounded by the harmful context bias to infer completely wrong emotional states.

\begin{figure}[t]
  \centering
  \includegraphics[width=\linewidth]{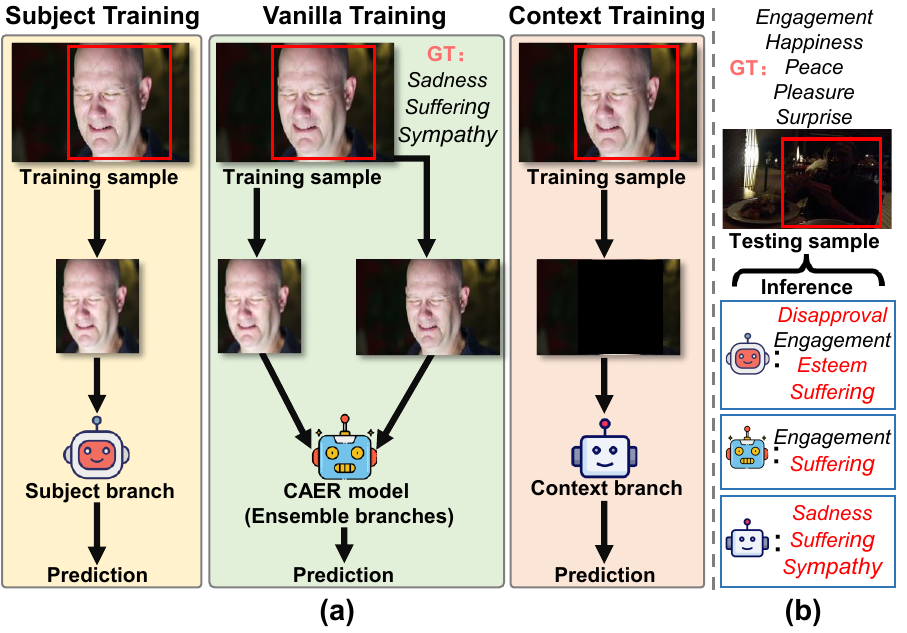}
  \caption{We conduct toy experiments to show the effects of context semantics. The indirect effect of the good context prior follows ensemble branches, narrowing the emotion candidate space. The bad direct effect follows the context branch, causing pure bias.
  }
  \label{intro2}
\vspace{-3pt}
\end{figure}

A straightforward solution is to conduct a randomized controlled trial by collecting images with all emotion annotations in all contexts. This manner is viewed as an approximate intervention for biased training. However, the current CAER debiasing effort~\cite{yang2023context} is sub-optimal since the predefined intervention fails to decouple good and bad context semantics. We argue that context semantics consists of the good prior and the bad bias. The toy experiments are performed to verify this insight. Specifically, we train on the EMOTIC dataset separately using the subject branch, the ensemble branches, and the context branch of a CAER baseline~\cite{kosti2017emotion} in \Cref{intro2}\textcolor{red}{a}. Recognized subjects in samples during context training are masked to capture the direct context effect. Observing the testing results in \Cref{intro2}\textcolor{red}{b}, the context prior in ensemble learning as the valuable indirect effect helps the model filter out unnecessary candidates (\ie, removing the ``\textit{Disapproval}'' and ``\textit{Esteem}'' categories) compared to the subject branch. Conversely, the harmful bias as the direct context effect in the context branch builds a misleading mapping between dim contexts and negative emotions during training, causing biased predictions.

To disentangle the two effects in context semantics and achieve more appropriate context debiasing, we propose a unified \underline{c}ounterfactua\underline{l} \underline{e}motion in\underline{f}erence (CLEF) framework from a causality perspective. CLEF focuses on assisting existing CAER methods to mitigate the context bias and breakthrough performance bottlenecks in a model-agnostic manner rather than beating them. Specifically, we first formulate a generalized causal graph to investigate causal relationships among variables in the CAER task. Along the causal graph, CLEF estimates the direct context effect caused by the harmful bias through a non-intrusive context branch during the training phase.
Meanwhile, the valuable indirect effect of the context prior in ensemble representations of subjects and contexts is calculated following the vanilla CAER model. In the inference phase, we subtract the direct context effect from the total causal effect by depicting a counterfactual scenario to exclude bias interference. This scenario is described as follows: 

\noindent \textbf{Counterfactual CAER:} \textit{What would the prediction be, if the model only sees the confounded context and does not perform inference via vanilla ensemble representations?}
Intuitively, ensemble representations in the counterfactual outcome are blocked in the no-treatment condition. As such, the model performs biased emotion estimation relying only on spurious correlations caused by the pure context bias, which results similarly to the predictions of the context branch in \Cref{intro2}\textcolor{red}{b}. By comparing factual and counterfactual outcomes, CLEF empowers the model to make unbiased predictions using the debiased causal effect.
The main contributions are summarized as follows: 
\begin{itemize}
\item We are the first to embrace counterfactual thinking to investigate causal effects in the CAER task and reveal that the context bias as the adverse direct causal effect misleads the models to produce spurious prediction shortcuts.
\item We devise CLEF, a model-agnostic CAER debiasing framework that facilitates existing methods to capture valuable causal relationships and mitigate the harmful bias in context semantics through counterfactual inference. CLEF can be readily adapted to state-of-the-art (SOTA) methods with different structures, bringing consistent and significant performance gains.
\item Extensive experiments are conducted on several large-scale CAER datasets. Comprehensive analyses show the broad applicability and effectiveness of our framework.
\end{itemize}

\section{Related Work}
\noindent \textbf{Context-Aware Emotion Recognition.}
Benefiting from advances in deep learning algorithms~\cite{wang2023sampling,wang2023adversarial,wang2023model,liu2023amp,liu2022learning,liu2023generalized,chen2024miss,wang2021tsa,wang2022spacenet,wang2023cpr,yang2023what2comm,yang2023spatio,yang2023how2comm},
traditional emotion recognition typically infers emotional states from subject-oriented attributes, such as facial expressions~\cite{du2021learning,li2020deep}, body postures~\cite{bhattacharya2020step,yang2023aide}, and acoustic behaviours~\cite{lian2021ctnet,mittal2020m3er}. However, these efforts are potentially vulnerable in practical applications since subject characteristics in uncontrolled environments are usually indistinguishable, leading to severe performance deterioration. Recently, a pioneering work~\cite{kosti2017emotion} inspired by psychological research~\cite{barrett2011context} has advocated extracting complementary emotional clues from rich contexts, called context-aware emotion recognition (CAER).  Kosti \etal \cite{kosti2019context} begin by utilizing a two-stream convolutional neural network (CNN) to capture effective semantics from subject-related regions and global contexts of complete images. The implementation is similar to the ensemble branch training in \Cref{intro2}\textcolor{red}{a}.
After that, most CAER methods~\cite{lee2019context,mittal2020emoticon,mittal2021multimodal,zhang2019context,li2021sequential,li2021human,hoang2021context,zeng2020graph,gao2021graph,yang2022emotion,wang2022context,chen2023incorporating,de2023high} follow an ensemble learning pattern: \textbf{i)} extracting unimodal/multimodal features from subject attributes; \textbf{ii)} learning emotionally relevant features from created contexts based on different definitions; and \textbf{iii)} producing ensemble representations for emotion predictions via fusion mechanisms. 
For instance, Yang \etal ~\cite{yang2022emotion} discretize the context into scenes, agent dynamics, and agent-object interactions, using customized components to learn complementary contextual information.  
Despite achievements, they invariably suffer from performance bottlenecks due to spurious correlations caused by the context bias.

\noindent \textbf{Causal Inference.}
Causal inference~\cite{glymour2016causal} is first extensively used in economics~\cite{varian2016causal} and psychology~\cite{foster2010causal} as a scientific theory that seeks causal relationships among variables.
The investigation of event causality generally follows two directions: intervention and counterfactuals.
Intervention~\cite{pearl2009causal} aims to actively manipulate the probability distributions of variables to obtain unbiased estimations or discover confounder effects.  Counterfactuals~\cite{pearl2009causality} typically utilize distinct treatment conditions to imagine outcomes that are contrary to factual determinations, empowering systems to reason and think like humans. In recent years, several learning-based approaches have attempted to introduce causal inference in diverse fields to pursue desired model effects and exclude the toxicity of spurious shortcuts, including scene graph generation~\cite{tang2020unbiased}, visual dialogue~\cite{qi2020two,niu2021counterfactual}, image recognition~\cite{wang2020visual,chalupka2014visual,lopez2017discovering}, and adversarial learning~\cite{kalainathan2018sam,kocaoglu2017causalgan}.
The CAER debiasing effort~\cite{yang2023context} most relevant to our work utilizes a predefined dictionary to approximate interventions and adopts memory-query operations to mitigate the bias dilemma. Nevertheless, the predefined-level intervention fails to capture pure bias effects in the context semantics, causing a sub-optimal solution. Inspired by~\cite{niu2021counterfactual}, we remove the adverse context effect by empowering models with the debiasing ability of twice-thinking through counterfactual causality, which is fundamentally different in design philosophy and methodology.

\begin{figure}[t]
  \centering
  \includegraphics[width=0.85\linewidth]{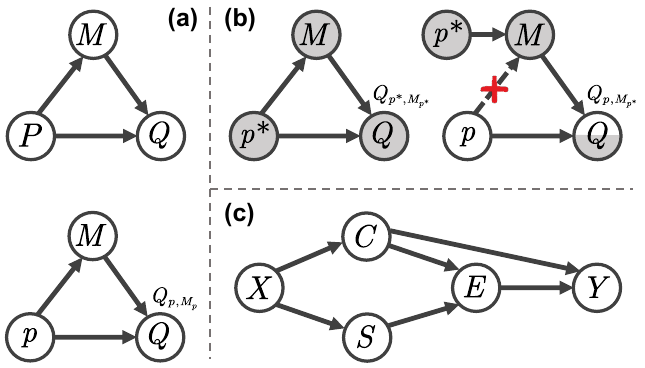}
  \caption{(a) Examples of a causal graph where nodes represent variables and arrows represent causal effects. (b) Examples of counterfactual notations. (c) The proposed CAER causal graph.
  }
  \label{graph}
\vspace{-6pt}
\end{figure}

\section{Preliminaries}
\label{sec3}
Before starting, we first introduce the concepts and notations related to causal inference to facilitate a better understanding of our framework and philosophy. 

\noindent \textbf{Causal graph} is a highly generalized analytical tool to reveal causal dependencies among variables.
It usually follows the structured causal mode~\cite{pearl2000models} defined as a directed acyclic graph $\mathcal{G} = \{ \mathcal{V}, \mathcal{E} \}$, where $\mathcal{V}$ stands for a set of variables and $\mathcal{E}$ implies the corresponding causal effects. A causal graph example with three variables is intuitively displayed in \Cref{graph}\textcolor{red}{a}. Here, we represent a random variable as a capital letter (\eg, $P$), and denote its observed value as a lowercase letter (\eg, $p$).
The causality from cause $P$ to effect $Q$ is reflected in two parts: the direct effect follows the causal link $ P \rightarrow Q$, and the indirect effect follows the link $P \rightarrow M \rightarrow Q$ through the mediator variable $M$.

\noindent \textbf{Counterfactual inference} endows the models with the ability to depict counterfactual outcomes in factual observations through different treatment conditions~\cite{pearl2009causality}.
In the factual outcome, the value of $Q$ would be formalized under the conditions that $P$ is set to $p$ and $M$ is set to $m$:
\begin{equation}
 \begin{split}
    Q_{p,m} = Q(P=p, M= m),\\
     m = M_p = M(P= p). \\
 \end{split}
 \end{equation}
Counterfactual outcomes can be obtained by exerting distinct treatments on the value of $P$. As shown in \Cref{graph}\textcolor{red}{b}, when $P$ is set to $p^*$, and the descendant $M$ is changed, we have $Q_{p^*,M_{p^*}} = Q(P=p^*, M_{p^*}= M(P=p^*))$. Similarly, $Q_{p, M_{p^*}}$ reflects the counterfactual situation where $P=p$ and $M$ is set to the value when $P=p^*$.

\begin{figure*}[t]
  \centering
  \includegraphics[width=0.95\textwidth]{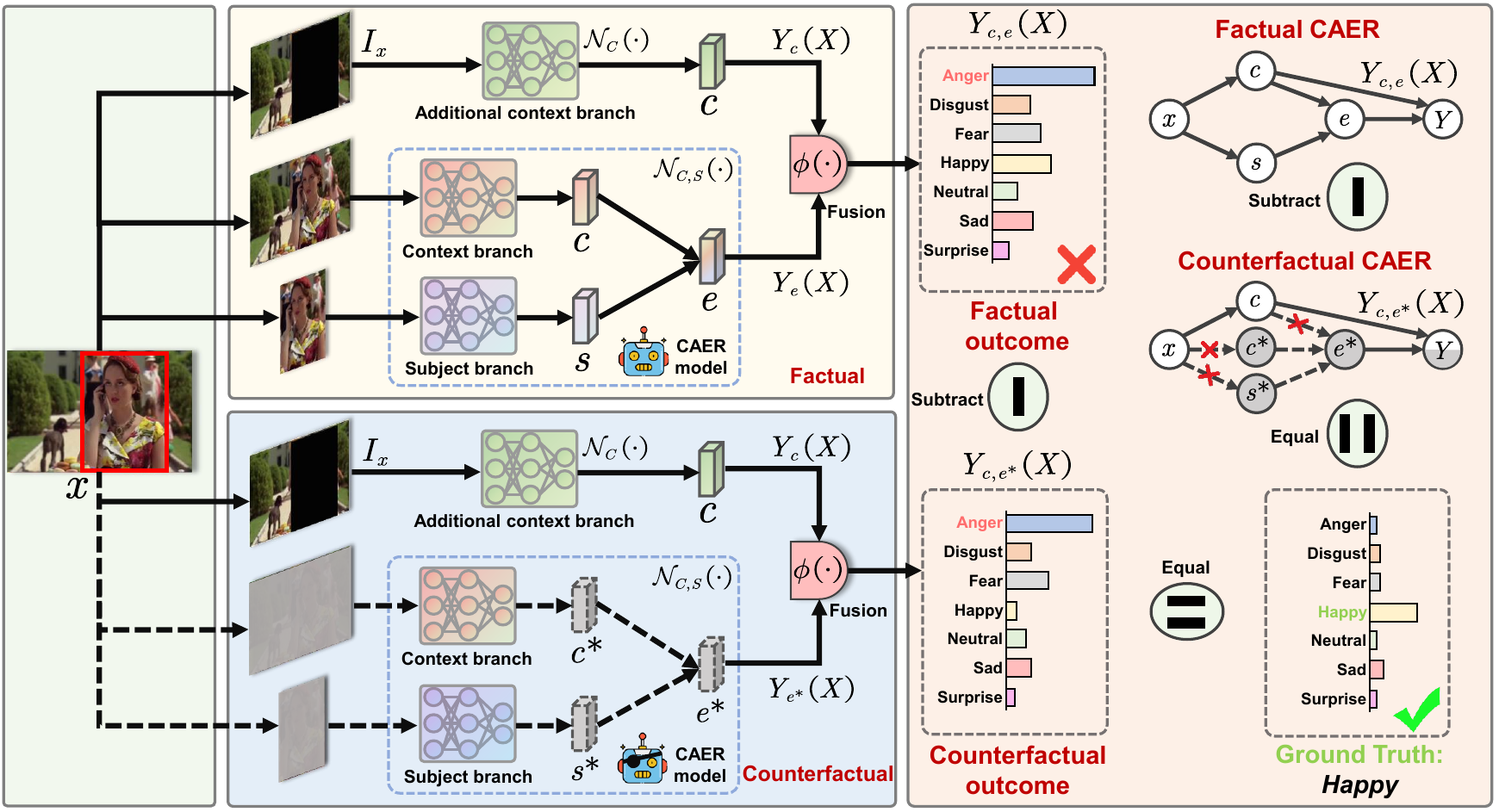}
  \caption{High-level overview of the proposed CLEF framework implementation. In addition to the vanilla CAER model, we introduce an additional context branch in a non-intrusive manner to capture the pure context bias
  as the direct context effect. By comparing factual and counterfactual outcomes, our framework effectively mitigates the interference of the harmful bias and achieves debiased emotion inference.
 }
  \label{arc}
  \vspace{-0.3cm}
\end{figure*}

\noindent \textbf{Causal effects} reveal the difference between two corresponding outcomes when the value of the reference variable changes. Let $P=p$ denote the treated condition and $P=p^*$ represent the invisible counterfactual condition.
According to the causal theory~\cite{pearl2014interpretation}, The Total Effect (TE) of treatment $P=p$ on $Q$ by comparing the two hypothetical outcomes is formulated as:
 \begin{equation}
\text{TE} = Q_{p, M_{p}}- Q_{p^*, M_{p^*}}.
 \end{equation}
TE can be disentangled into the Natural Direct Effect (NDE) and the Total Indirect Effect (TIE)~\cite{glymour2016causal}.
NDE reflects the effect of $P=p$ on $Q$ following the direct link $ P \rightarrow Q$, and excluding the indirect effect along link $P \rightarrow M \rightarrow Q$ due to $M$ is set to the value when $P$ had been $p^*$. It reveals the response of $Q$ when $P$ converts from $p$ to $p^*$:
 \begin{equation}
\text{NDE} = Q_{p, M_{p^*}}- Q_{p^*, M_{p^*}}.
 \end{equation}
In this case, TIE is calculated by directly subtracting NDE from TE, which is employed to measure the unbiased prediction results in our framework:
 \begin{equation}
\text{TIE} = \text{TE}- \text{NDE} = Q_{p, M_{p}} - Q_{p, M_{p^*}}.
 \end{equation}

\section{The proposed CLEF Framework}
\subsection{Cause-Effect Look at CAER}
As shown in \Cref{graph}\textcolor{red}{c}, there are five variables in the proposed CAER causal graph, including input images $X$, subject features $S$, context features $C$, ensemble representations $E$, and emotion predictions $Y$.
Note that our causal graph has broad applicability and generality since it follows most CAER modelling paradigms.

\noindent \textbf{Link} $ \bm{X} \rightarrow \bm{C} \rightarrow \bm{Y}$ reflects the shortcut between the original inputs $X$ and the model predictions $Y$ through the harmful bias in the context features $C$. The adverse direct effect of the mediator $C$ is obtained via a non-invasive branch of context modelling, which captures spurious correlations between context-specific semantics and emotion labels.
Taking \Cref{intro2}\textcolor{red}{b} as an example, the context branch learns the undesired mapping between dim contexts and negative emotions during training.

\noindent \textbf{Link} $ \bm{C} \leftarrow \bm{X} \rightarrow \bm{S}$ portrays the total context and subject representations extracted from $X$ via the corresponding encoders in vanilla CAER models. Based on design differences in distinct methods, $C$ and $S$ may come from a single feature or an aggregation of multiple sub-features. For instance, $S$ is obtained from global body attributes and joint face-pose information in models~\cite{kosti2017emotion} and \cite{mittal2020emoticon}, respectively.

\noindent \textbf{Link} $ \bm{C/S} \rightarrow \bm{E} \rightarrow \bm{Y}$ captures the indirect causal effect of $C$ and $S$ on the model predictions $Y$ through the ensemble representations $E$. The mediator $E$ is obtained depending on the feature integration mechanisms of different vanilla methods, such as feature concatenation~\cite{kosti2017emotion} or attention fusion~\cite{lee2019context}. In particular, $C$ provides the valuable context prior along the good causal link $ \ C \rightarrow E \rightarrow Y$, which gives favorable estimations of potential emotional states when the subjects' characteristics are indistinguishable.

\subsection{Counterfactual Inference}
Our design philosophy is to mitigate the interference of the harmful context bias on model predictions by excluding the biased direct effect along the link $  X \rightarrow C \rightarrow Y$. Following the notations on causal effects in \Cref{sec3}, the causality in the factual scenarios is formulated as follows:
 \begin{equation}
Y_{c,e}(X) = Y(C=c, E_{c,s} = E(C=c, S=s)|X).
 \end{equation}
$Y_{c,e}(X)$ reflects confounded emotion predictions because it suffers from the detrimental direct effect of $C$, \textit{i.e.}, the pure context bias. To disentangle distinct causal effects in the context semantics, we calculate the Total Effect (TE) of $C = c$ and $S = s$, which is expressed as follows:
 \begin{equation}
\text{TE} = Y_{c,e}(X) - Y_{c^*,e^*}(X).
 \end{equation}
Here, $c^*$ and $e^*$ represent the non-treatment conditions for observed values of $C$ and $E$, where $c$ and $s$ leading to $e$ are not given.
Immediately, we estimate the Natural Direct Effect (NDE) for the harmful bias in context semantics:
 \begin{equation}
\text{NDE} = Y_{c,e^*}(X) - Y_{c^*,e^*}(X).
 \end{equation}
$Y_{c,e^*}(X)$ describes a counterfactual outcome where $C$ is set to $c$ and $E$ would be imagined to be $e^*$ when $C$ had been $c^*$ and $S$ had been $s^*$. The causal notation is expressed as:
 \begin{equation}
Y_{c,e^*}(X) = Y(C=c, E_{c^*,s^*} = E(C=c^*, S= s^*)|X).
 \end{equation}
Since the indirect causal effect of ensemble representations $E$ on the link $  X \rightarrow C/S \rightarrow E \rightarrow Y$ is blocked, the model can only perform biased predictions by relying on the direct context effect in the link $  X \rightarrow C \rightarrow Y$ that causes spurious correlations. To exclude the explicitly captured context bias in NDE, we subtract NDE from TE to estimate Total Indirect Effect (TIE):
 \begin{equation}
\text{TIE} = Y_{c,e}(X) - Y_{c,e^*}(X).
\label{eq9}
 \end{equation}
We employ the reliable TIE as the unbiased prediction in the inference phase.

\subsection{Implementation Instantiation}

\noindent \textbf{Framework Structure.} From \Cref{arc}, CLEF's predictions consist of two parts: the prediction $Y_{c}(X) = \mathcal{N}_{C}(c|x)$ of the additional context branch (\ie, $ X \rightarrow C \rightarrow Y$) and $Y_{e}(X) = \mathcal{N}_{C,S}(c,s|x)$ of the vanilla CAER model (\ie, $  X \rightarrow C/S \rightarrow E \rightarrow Y$). The context branch is instantiated as a simple neural network $\mathcal{N}_{C}(\cdot)$ (\eg, ResNet~\cite{he2016deep}) to receive context images with masked recognized subjects. The masking operation forces the network to focus on pure context semantics for estimating the direct effect. 
For a given input $x$, its corresponding context image $I_x$ is expressed as:
\begin{equation}
    I_{x}= \begin{cases}x(i, j) & \text { if } x(i, j) \notin \text {bbox}_{\text {subject }}, \\ 0 & \text { otherwise }, \end{cases}
\label{bbx}
\end{equation}
where $\text{bbox}_\text{subject}$ means the bounding box of the subject. $\mathcal{N}_{C,S}(\cdot)$ denotes any CAER model based on their specific mechanisms to learn ensemble representations $e$ from $c$ and $s$ for prediction. 
Subsequently, a pragmatic fusion strategy $\phi(\cdot)$ is introduced to obtain the final score $Y_{c,e}(X)$:
\begin{equation}
   Y_{c,e}(X) = \phi(Y_{c}(X), Y_{e}(X)) = \text{log} \sigma(Y_{c}(X) + Y_{e}(X)),
\label{eq11}
\end{equation}
where $\sigma$ is the sigmoid activation.

\noindent \textbf{Training Procedure.} As a universal framework, we take the multi-class classification task in \Cref{arc} as an example to adopt the cross-entropy loss $\mathcal{CE}(\cdot)$ as the optimization objective. The task-specific losses for $Y_{c,e}(X)$ and $Y_{c,e^*}(X)$ are as follows:
\begin{equation}
   \mathcal{L}_{task} = \mathcal{CE}(Y_{c,e}(X), y) + \mathcal{CE}(Y_{c,e^*}(X), y),
\end{equation}
where $y$ means the ground truth.
Since neural models cannot handle no-treatment conditions where the inputs are void, we devise a trainable parameter initialized by the uniform distribution in practice to represent the imagined $Y_{e^*}(X)$, which is shared by all samples. The design intuition is that the uniform distribution ensures a safe estimation for NDE, which is justified in subsequent ablation studies. To avoid inappropriate $Y_{e^*}(X)$ that potentially causes TIE to be dominated by TE or NDE, we employ the Kullback-Leibler divergence $\mathcal{KL}(\cdot)$ to regularize the difference between $Y_{c,e^*}(X)$ and $Y_{c,e}(X)$ to estimate $Y_{e^*}(X)$:
\begin{equation} 
  \mathcal{L}_{kl} = \mathcal{KL}(Y_{c,e^*}(X), Y_{c,e}(X)).
\end{equation}
The final loss is expressed as:
\begin{equation}
  \mathcal{L}_{fin} =  \sum_{(c,s,y) \in \mathcal{D}}^{} \mathcal{L}_{task} + \mathcal{L}_{kl}.
\end{equation}

\noindent \textbf{Inference Procedure.} According to Eq.\,(\ref{eq9}), the debiased prediction is performed as follows:
\begin{equation}
  \text{TIE} =  \phi(Y_{c}(X), Y_{e}(X)) - \phi(Y_{c}(X), Y_{e*}(X)).
\end{equation}

\section{Experiments}

\begin{table*}[t]
\centering
\caption{ Quantitative results of CLEF-based methods for each emotion category on the EMOTIC dataset. We report the average precision of each category to provide comprehensive comparison experiments. The improved results are marked in \textbf{bold}.}
\vspace{-0.2cm}
\resizebox{\linewidth}{!}{%
\begin{tabular}{cl|cccccccccc}
\toprule
\multicolumn{2}{c|}{\multirow{2}{*}{\textbf{Category}}} & \multirow{2}{*}{EMOT-Net~\cite{kosti2019context}} & \multirow{2}{*}{\begin{tabular}[c]{@{}c@{}}EMOT-Net\\ + \textbf{CLEF}\end{tabular}} & \multirow{2}{*}{CAER-Net~\cite{lee2019context}} & \multirow{2}{*}{\begin{tabular}[c]{@{}c@{}}CAER-Net\\ + \textbf{CLEF}\end{tabular}} & \multirow{2}{*}{GNN-CNN~\cite{zhang2019context}} & \multirow{2}{*}{\begin{tabular}[c]{@{}c@{}}GNN-CNN\\ + \textbf{CLEF}\end{tabular}} & \multirow{2}{*}{CD-Net~\cite{wang2022context}} & \multirow{2}{*}{\begin{tabular}[c]{@{}c@{}}CD-Net\\ + \textbf{CLEF}\end{tabular}} & \multirow{2}{*}{EmotiCon~\cite{mittal2020emoticon}} & \multirow{2}{*}{\begin{tabular}[c]{@{}c@{}}EmotiCon\\ + \textbf{CLEF}\end{tabular}} \\
\multicolumn{2}{c|}{}                                   &                                    &                                                                                    &                                    &                                                                                    &                                   &                                                                                   &                                  &                                                                                  &                                    &                                                                                    \\ \midrule
\multicolumn{2}{c|}{Affection}                          & 26.47                              & 35.28                                                                              & 22.36                              & 28.62                                                                              & 47.52                             & 61.84                                                                             & 28.44                            & 35.51                                                                            & 38.55                              & 43.72                                                                              \\
\multicolumn{2}{c|}{Anger}                              & 11.24                              & 11.76                                                                              & 12.88                              & 14.01                                                                              & 11.27                             & 16.37                                                                             & 12.12                            & 14.6                                                                             & 14.69                              & 17.09                                                                              \\
\multicolumn{2}{c|}{Annoyance}                          & 15.26                              & 17.46                                                                              & 14.42                              & 12.85                                                                              & 12.33                             & 11.08                                                                             & 19.71                            & 16.94                                                                            & 24.68                              & 25.40                                                                               \\
\multicolumn{2}{c|}{Anticipation}                       & 57.31                              & 94.29                                                                              & 52.85                              & 82.27                                                                              & 63.20                              & 93.25                                                                             & 57.65                            & 89.05                                                                            & 60.73                              & 92.24                                                                              \\
\multicolumn{2}{c|}{Aversion}                           & 7.44                               & 13.14                                                                              & 3.26                               & 10.23                                                                              & 6.81                              & 10.30                                                                              & 9.94                             & 16.83                                                                            & 11.33                              & 15.51                                                                              \\
\multicolumn{2}{c|}{Confidence}                         & 80.33                              & 74.48                                                                              & 72.68                              & 73.18                                                                              & 74.83                             & 69.02                                                                             & 69.26                            & 73.11                                                                            & 68.12                              & 65.90                                                                               \\
\multicolumn{2}{c|}{Disapproval}                        & 16.14                              & 19.73                                                                              & 15.37                              & 17.04                                                                              & 12.64                             & 15.16                                                                             & 22.78                            & 27.45                                                                            & 18.55                              & 21.47                                                                              \\
\multicolumn{2}{c|}{Disconnection}                      & 20.64                              & 30.66                                                                              & 22.01                              & 24.76                                                                              & 23.17                             & 28.35                                                                             & 27.55                            & 31.70                                                                            & 28.73                              & 33.31                                                                              \\
\multicolumn{2}{c|}{Disquietment}                       & 19.57                              & 19.73                                                                              & 10.84                              & 13.47                                                                              & 17.66                             & 20.11                                                                             & 21.04                            & 23.37                                                                            & 22.14                              & 24.56                                                                              \\
\multicolumn{2}{c|}{Doubt/Confusion}                    & 31.88                              & 19.81                                                                              & 26.07                              & 22.15                                                                              & 19.67                             & 16.57                                                                             & 24.23                            & 19.55                                                                            & 38.43                              & 32.87                                                                              \\
\multicolumn{2}{c|}{Embarrassment}                      & 3.05                               & 6.53                                                                               & 1.88                               & 5.31                                                                               & 1.58                              & 4.08                                                                              & 4.50                              & 7.24                                                                             & 10.31                              & 12.98                                                                              \\
\multicolumn{2}{c|}{Engagement}                         & 86.69                              & 97.39                                                                              & 73.71                              & 90.46                                                                              & 87.31                             & 92.88                                                                             & 85.32                            & 94.38                                                                            & 86.23                              & 92.75                                                                              \\
\multicolumn{2}{c|}{Esteem}                             & 17.86                              & 22.30                                                                               & 15.38                              & 17.91                                                                              & 12.05                             & 18.69                                                                             & 18.66                            & 23.01                                                                            & 25.75                              & 29.13                                                                              \\
\multicolumn{2}{c|}{Excitement}                         & 78.05                              & 73.36                                                                              & 70.42                              & 63.01                                                                              & 72.68                             & 65.21                                                                             & 70.07                            & 60.42                                                                            & 80.75                              & 72.64                                                                              \\
\multicolumn{2}{c|}{Fatigue}                            & 8.87                               & 10.34                                                                              & 6.29                               & 8.66                                                                               & 12.93                             & 17.67                                                                             & 11.56                            & 14.67                                                                            & 19.35                              & 22.34                                                                              \\
\multicolumn{2}{c|}{Fear}                               & 15.70                               & 8.46                                                                               & 7.47                               & 10.12                                                                              & 6.15                              & 10.34                                                                             & 10.38                            & 11.23                                                                            & 16.99                              & 18.71                                                                              \\
\multicolumn{2}{c|}{Happiness}                          & 58.92                              & 77.89                                                                              & 53.73                              & 72.37                                                                              & 72.90                              & 81.79                                                                             & 68.46                            & 84.24                                                                            & 80.45                              & 87.06                                                                              \\
\multicolumn{2}{c|}{Pain}                               & 9.46                               & 13.97                                                                              & 8.16                               & 10.32                                                                              & 8.22                              & 11.94                                                                             & 13.82                            & 16.44                                                                            & 14.68                              & 15.45                                                                              \\
\multicolumn{2}{c|}{Peace}                              & 22.35                              & 23.23                                                                              & 19.55                              & 20.05                                                                              & 30.68                             & 31.56                                                                             & 28.18                            & 26.05                                                                            & 35.72                              & 35.96                                                                              \\
\multicolumn{2}{c|}{Pleasure}                           & 46.72                              & 45.92                                                                              & 34.12                              & 34.46                                                                              & 48.37                             & 51.73                                                                             & 47.64                            & 50.92                                                                            & 67.31                              & 68.42                                                                              \\
\multicolumn{2}{c|}{Sadness}                            & 18.69                              & 27.19                                                                              & 17.75                              & 23.06                                                                              & 23.90                              & 33.28                                                                             & 32.99                            & 37.43                                                                            & 40.26                              & 45.25                                                                              \\
\multicolumn{2}{c|}{Sensitivity}                        & 9.05                               & 7.84                                                                               & 6.94                               & 8.12                                                                               & 4.74                              & 5.14                                                                              & 7.21                             & 10.70                                                                             & 13.94                              & 15.07                                                                              \\
\multicolumn{2}{c|}{Suffering}                          & 17.67                              & 18.05                                                                              & 14.85                              & 15.63                                                                              & 23.71                             & 25.60                                                                              & 35.19                            & 30.85                                                                            & 48.05                              & 43.16                                                                              \\
\multicolumn{2}{c|}{Surprise}                           & 22.38                              & 12.27                                                                              & 17.46                              & 14.70                                                                               & 8.44                              & 6.01                                                                              & 7.42                             & 7.21                                                                             & 19.60                               & 20.18                                                                              \\
\multicolumn{2}{c|}{Sympathy}                           & 15.23                              & 30.15                                                                              & 14.89                              & 15.53                                                                              & 19.45                             & 25.13                                                                             & 10.33                            & 13.66                                                                            & 16.74                              & 20.64                                                                              \\
\multicolumn{2}{c|}{Yearning}                           & 9.22                               & 12.13                                                                              & 4.84                               & 5.16                                                                               & 9.86                              & 13.64                                                                             & 6.24                             & 8.63                                                                             & 15.08                              & 17.39                                                                              \\ \midrule
\multicolumn{2}{c|}{\textbf{mAP} (\%)}                       & 27.93                              & \textbf{31.67}                                                                     & 23.85                              & \textbf{27.44}                                                                     & 28.16                             & \textbf{32.18}                                                                    & 28.87                            & \textbf{32.51}                                                                   & 35.28                              & \textbf{38.05}                                                                     \\ \bottomrule
\end{tabular}
}
\label{tab1}
\vspace{-4pt}
\end{table*}

\subsection{Datasets and Evaluation Metrics}
Experiments are conducted on two large-scale image-based CAER datasets, including EMOTIC~\cite{kosti2019context} and CAER-S~\cite{lee2019context}. 
\textbf{EMOTIC} is the first benchmark to support emotion recognition in real-world contexts, which has 23,571 images of 34,320 annotated subjects.
All samples are collected from non-controlled environments to provide rich context resources. Each recognized subject is annotated with 26 discrete emotion categories and body bounding box information. The dataset is partitioned into 70\% samples for training, 10\% samples for validation, and 20\% samples for testing.
\textbf{CAER-S} consists of 70k static images extracted from video clips. These images record 7 emotional states of different subjects in various context scenarios from 79 TV shows. 
The data samples are randomly divided into training, validation, and testing sets in the ratio of 7:1:2.
We utilize the standard mean Average Precision (mAP) and classification accuracy to evaluate the results on the EMOTIC and CAER-S datasets, respectively.

\subsection{Model Zoo}
We evaluate the effectiveness of the proposed CLEF using five representative methods, which have completely different network structures and contextual modelling paradigms. Concretely, \textbf{EMOT-Net}~\cite{kosti2017emotion} is a two-stream classical CNN model where one stream extracts human features from body regions, and the other captures global context semantics.  \textbf{CAER-Net}~\cite{lee2019context} extracts subject attributes from faces and uses the images after hiding faces as background contexts. 
\textbf{GNN-CNN}~\cite{zhang2019context} utilizes the graph neural network (GNN) to integrate emotion-related objects in contexts and distills subject information with a VGG-16~\cite{simonyan2014very}. \textbf{CD-Net}~\cite{wang2022context} designs a tube-transformer to perform fine-grained interactions from facial, bodily, and contextual features. \textbf{EmotiCon}~\cite{mittal2020emoticon} employs attention and depth maps to model context representations. Subject-relevant features are extracted from facial expressions and body postures.

\subsection{Implementation Details}

\begin{figure*}[t]
  \centering
  \includegraphics[width=0.9\textwidth]{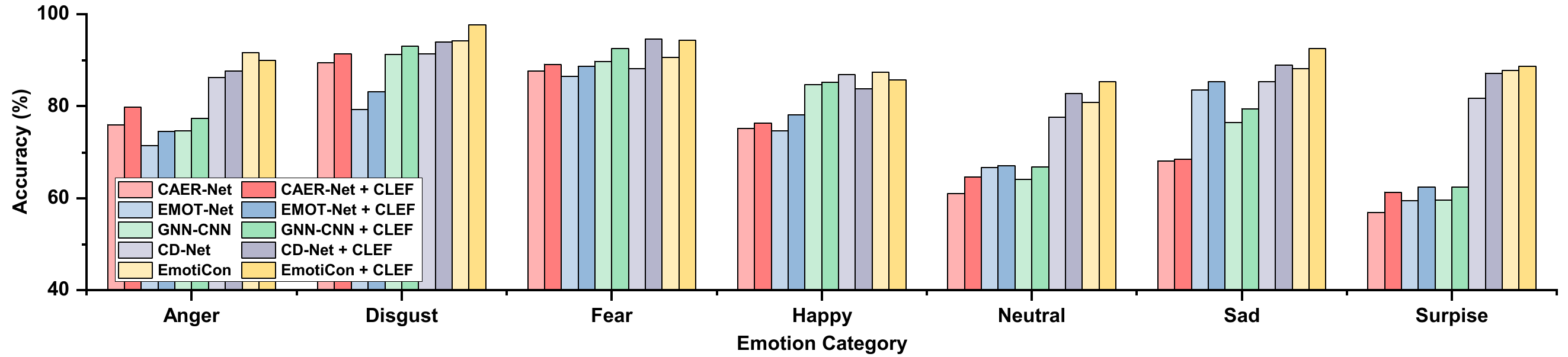}
  \caption{Emotion classification accuracy (\%) for each category of different CLEF-based methods on the CAER-S dataset.
 }
  \label{fig-caer}
  \vspace{-0.3cm}
\end{figure*}

We use a ResNet-152~\cite{he2016deep} pre-trained on the Places365~\cite{zhou2017places} dataset to parameterize the non-invasive context branch in CLEF. The output of the last linear layer is replaced to produce task-specific numbers of neurons for predictions. Rich scene attributes in Places365 provide proper semantics for distilling the context bias. In addition to the annotated EMOTIC, we employ the Faster R-CNN~\cite{ren2015faster} to detect bounding boxes of recognized subjects in CAER-S.
Immediately, the context images $I_x$ are obtained by masking the target subjects in samples based on the corresponding bounding boxes. For a fair comparison, the five selected CAER methods are reproduced via the PyTorch toolbox~\cite{paszke2017automatic} following their reported training settings, including the optimizer, loss function, learning rate strategy, etc.
All models are implemented on NVIDIA Tesla V100 GPUs.

\subsection{Comparison with State-of-the-art Methods}
We compare the five CLEF-based methods with existing SOTA models, including HLCR~\cite{de2023high}, TEKG~\cite{chen2023incorporating}, RRLA~\cite{li2021human}, VRD~\cite{hoang2021context}, SIB-Net~\cite{li2021sequential}, MCA~\cite{yang2022emotion}, and GRERN~\cite{gao2021graph}.

\noindent \textbf{Quantitative Results on the EMOTIC.}
\Cref{tab1} shows the Average Precision (AP) of the vanilla methods and their counterparts in the CLEF framework for each emotion category. We have the following critical observations. \textbf{i)} CLEF significantly improves the performance of all models in most categories. For instance, CLEF yields average gains of  8.33\% and 6.52\% on the AP scores for ``\textit{Affection}'' and ``\textit{Sadness}'', reflecting positivity and negativity, respectively. 
\textbf{ii)} Our framework favorably improves several categories heavily confounded by the harmful context bias due to uneven distributions of emotional states across distinct contexts. 
For example, the CLEF-based models improve the AP scores for ``\textit{Engagement}'' and ``\textit{Happiness}'' categories to 90.46\%$\sim$97.39\%  and 72.37\%$\sim$87.06\%, outperforming the results in the vanilla baselines by large margins.

\begin{table}[t]
\setlength{\tabcolsep}{35pt}
\centering
\caption{Quantitative results of different models and CLEF-based methods on the EMOTIC dataset.  $\uparrow$ represents the improvement of the CLEF-based version over the vanilla method.}
\vspace{-0.2cm}
\resizebox{0.85\linewidth}{!}{%
\begin{tabular}{c|c}
\toprule
\textbf{Methods}         & \textbf{mAP} (\%)            \\ \midrule
HLCR~\cite{de2023high}            & 30.02          \\
TEKG~\cite{chen2023incorporating}            & 31.36          \\
RRLA~\cite{li2021human}            & 32.41          \\
VRD~\cite{hoang2021context}             & 35.16          \\
SIB-Net~\cite{li2021sequential}         & 35.41          \\
MCA~\cite{yang2022emotion}             & 37.73          \\ \midrule
EMOT-Net~\cite{kosti2019context}        & 27.93          \\
EMOT-Net + \textbf{CLEF} & \textbf{31.67} (${\uparrow}$ \textcolor{pink}{3.74}) \\
CAER-Net~\cite{lee2019context}        & 23.85          \\
CAER-Net + \textbf{CLEF} & \textbf{27.44} (${\uparrow}$ \textcolor{pink}{3.59}) \\
GNN-CNN~\cite{zhang2019context}         & 28.16          \\
GNN-CNN + \textbf{CLEF}  & \textbf{32.18} (${\uparrow}$ \textcolor{pink}{4.02})\\
CD-Net~\cite{wang2022context}          & 28.87          \\
CD-Net + \textbf{CLEF}   & \textbf{32.51} (${\uparrow}$ \textcolor{pink}{3.64})\\
EmotiCon~\cite{mittal2020emoticon}        & 35.28          \\
EmotiCon + \textbf{CLEF} & \textbf{38.05} (${\uparrow}$ \textcolor{pink}{2.77})\\ \bottomrule
\end{tabular}
}
\label{tab2}
\vspace{-12pt}
\end{table}

\Cref{tab2} presents the comparison results with existing models regarding the mean AP (mAP) scores.
\textbf{i)} Thanks to CLEF's bias exclusion, the mAP scores of EMOT-Net, CAER-Net, GNN-CNN, CD-Net, and EmotiCon are consistently increased by 3.74\%, 3.59\%, 4.02\%, 3.64\%, and 2.77\%, respectively.  Among them, the most noticeable improvement in GNN-CNN is because the vanilla model more easily captures spurious \textit{context-emotion} correlations based on fine-grained context element exploration~\cite{zhang2019context}, leading to the better debiasing effect with CLEF.
\textbf{ii)} Compared to SIB-Net and MCA with complex module stacking~\cite{yang2022emotion} and massive parameters~\cite{li2021sequential}, the CLEF-based EmotiCon achieves the best performance with the mAP score of 38.05\% through efficient counterfactual inference.

\begin{table}[t]
\setlength{\tabcolsep}{35pt}
\centering
\caption{Quantitative results of different models and CLEF-based methods on the CAER-S dataset.}
\vspace{-0.2cm}
\resizebox{0.85\linewidth}{!}{%
\begin{tabular}{c|c}
\toprule
\textbf{Methods}           & \textbf{Accuracy} (\%)       \\ \midrule
Fine-tuned VGGNet~\cite{simonyan2014very} & 64.85          \\
Fine-tuned ResNet~\cite{he2016deep} & 68.46          \\
SIB-Net~\cite{li2021sequential}           & 74.56          \\
MCA~\cite{yang2022emotion}               & 79.57          \\
GRERN~\cite{gao2021graph}             & 81.31          \\
RRLA~\cite{li2021human}              & 84.82          \\
VRD~\cite{hoang2021context}               & 90.49          \\ \midrule
EMOT-Net~\cite{kosti2019context}          & 74.51          \\
EMOT-Net + \textbf{CLEF}   & \textbf{77.03} (${\uparrow}$ \textcolor{pink}{2.52})\\
CAER-Net~\cite{lee2019context}          & 73.47          \\
CAER-Net + \textbf{CLEF}   & \textbf{75.86} (${\uparrow}$ \textcolor{pink}{2.39})\\
GNN-CNN~\cite{zhang2019context}           & 77.21          \\
GNN-CNN + \textbf{CLEF}    & \textbf{79.53} (${\uparrow}$ \textcolor{pink}{2.32})\\
CD-Net~\cite{wang2022context}            & 85.33          \\
CD-Net + \textbf{CLEF}     & \textbf{88.41} (${\uparrow}$ \textcolor{pink}{3.08}) \\
EmotiCon~\cite{mittal2020emoticon}          & 88.65          \\
EmotiCon + \textbf{CLEF}   & \textbf{90.62} (${\uparrow}$ \textcolor{pink}{1.97})\\ \bottomrule
\end{tabular}
}
\label{tab:caer}
\vspace{-8pt}
\end{table}

\noindent \textbf{Quantitative Results on the CAER-S.}
\Cref{tab:caer} provides the evaluation results on the CAER-S dataset.
\textbf{i)} Evidently, CLEF consistently improves different baselines by decoupling and excluding the prediction bias of emotional states in the TV show contexts. Concretely, the overall accuracies of EMOT-Net, CAER-Net, GNN-CNN, CD-Net, and EmotiCon are improved by 2.52\%, 2.39\%, 2.32\%, 3.08\%, and 1.97\%, respectively.
\textbf{ii)} The gains of our framework on the CAER-S are slightly weaker than those on the EMOTIC. A reasonable explanation is that the EMOTIC contains richer context semantics than the CAER-S, such as scene elements and agent dynamics~\cite{kosti2019context}. As a result, CLEF more accurately estimates the adverse context effect and favorably removes its interference.
\textbf{iii)} Also, we find in \Cref{fig-caer} that the classification accuracies of most emotion categories across the five methods are improved appropriately.

\subsection{Ablation Studies}
In \Cref{abl}, we select the SOTA CD-Net and EmotiCon to perform thorough ablation studies on both datasets.

\begin{figure*}[t]
  \centering
  \includegraphics[width=\textwidth]{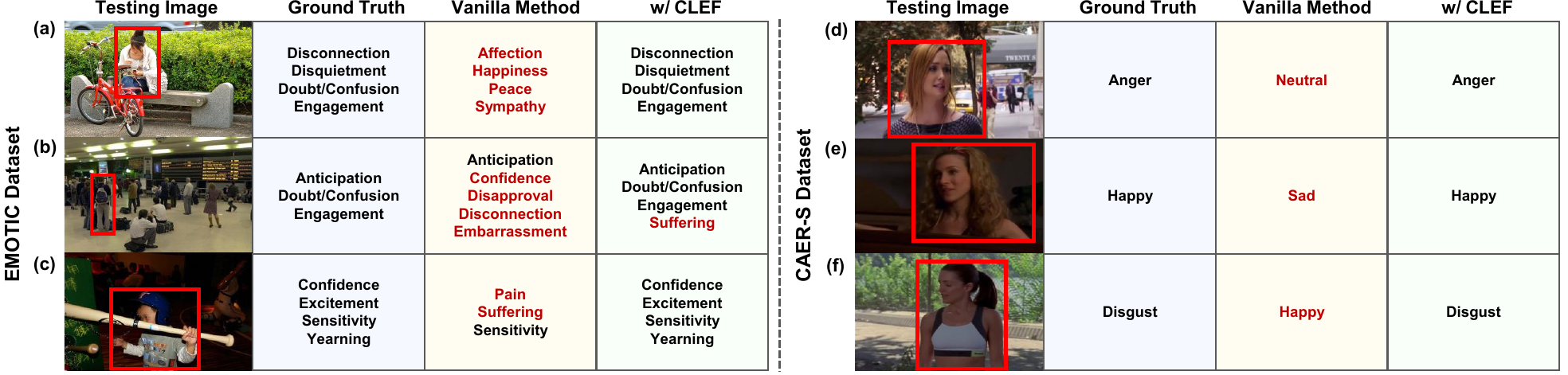}
  \caption{Qualitative results of the vanilla and CLEF-based CD-Net~\cite{wang2022context} on the EMOTIC and CAER-S datasets. Three testing sample images on each dataset are randomly selected. Incorrectly predicted categories are marked in \textcolor{red}{red}.
 }
  \label{vis}
  \vspace{-0.3cm}
\end{figure*}

\begin{table}[t]
\centering
\caption{Ablation study results on the EMOTIC and CAER-S datasets. ``ACB'' means the additional context branch. ``w/'' and ``w/o'' are short for the with and without, respectively.}
\vspace{-0.2cm}
\resizebox{\linewidth}{!}{%
\begin{tabular}{ccccc}
\toprule
\multicolumn{1}{c|}{\multirow{2}{*}{\textbf{Setting}}}     & \multicolumn{2}{c}{\textbf{EMOTIC}~\cite{kosti2019context}}      & \multicolumn{2}{c}{\textbf{CAER-S}~\cite{lee2019context}}      \\ \cline{2-5} 
\multicolumn{1}{c|}{}                             & CD-Net         & EmotiCon       & CD-Net         & EmotiCon       \\ \midrule
\multicolumn{1}{c|}{Vanilla Method}               & 28.87          & 35.28          & 85.33          & 88.65          \\ \midrule
\multicolumn{5}{c}{Necessity of  Framework Structure}                                                                \\ \midrule
\multicolumn{1}{c|}{+ CLEF}                       & \textbf{32.51} & \textbf{38.05} & \textbf{88.41} & \textbf{90.62} \\
\multicolumn{1}{c|}{w/o CAER Model}               & 19.64          & 19.64          & 62.87          & 62.87          \\
\multicolumn{1}{c|}{w/o ACB}                      & 28.55          & 35.43          & 85.54          & 88.28          \\
\multicolumn{1}{c|}{w/o $\mathcal{KL}(\cdot)$ Regularization}        & 32.26          & 37.44          & 88.09          & 90.36          \\ \midrule
\multicolumn{5}{c}{Rationality of  Context Modelling}                                                                 \\ \midrule
\multicolumn{1}{c|}{w/o Masking Operation}        & 31.38          & 36.95          & 87.68          & 89.85          \\
\multicolumn{1}{c|}{w/ ImageNet Pre-training}     & 30.74          & 36.62          & 87.35          & 89.27          \\
\multicolumn{1}{c|}{w/ ResNet-50~\cite{he2016deep}}                 & 31.45          & 37.54          & 87.83          & 90.04          \\
\multicolumn{1}{c|}{w/ VGG-16~\cite{simonyan2014very}}                    & 29.93          & 36.48          & 86.76          & 89.39          \\ \midrule
\multicolumn{5}{c}{Effectiveness of No-treatment Assumption}                                                        \\ \midrule
\multicolumn{1}{c|}{w/ Average Feature Embedding} & 27.85          & 33.18          & 83.06          & 85.67          \\
\multicolumn{1}{c|}{w/ Random Feature Embedding}  & 24.61          & 28.77          & 76.43          & 78.25          \\ \bottomrule
\end{tabular}
}
\label{abl}
\vspace{-6pt}
\end{table}

\noindent \textbf{Necessity of Framework Structure.}
\textbf{i)} When removing CAER models from CLEF, the significant performance deterioration suggests that the indirect causal effect in ensemble representations provides valuable emotion semantics.
\textbf{ii)} When the additional context branch (ACB) is excluded, CLEF degrades to a debiased pattern that is not context-conditional, treated as TE. TE's gains are inferior to TIE's since it reduces the general bias over the whole dataset rather than the specific context bias.
\textbf{iii)} Also, we find that the $\mathcal{KL}(\cdot)$ regularization is indispensable for estimating the proper $Y_{e^*}(X)$ and improving debiasing gains.

\noindent \textbf{Rationality of Context Modelling.}
\textbf{i)} We observe that performing the masking operation on target subjects in input images of ACB is essential for ensuring reliable capture of the context-oriented adverse direct effect.
\textbf{ii)} When the ResNet-152 pre-trained on Places365~\cite{zhou2017places} is replaced with the one pre-trained on ImageNet~\cite{deng2009imagenet} in ACB, the gain drops prove that scene-level semantics are more expressive than object-level semantics in reflecting the context bias. This makes sense since scene attributes usually contain diverse object concepts.
\textbf{iii)} Moreover, the improvements from CLEF gradually increase as more advanced pre-training backbones are used, which shows that our framework does not rely on a specific selection of instantiated networks.

\noindent \textbf{Effectiveness of No-treatment Assumption.}
We provide two alternatives regarding the no-treatment condition assumption, where random and average feature embeddings are obtained by the random initialization and the prior distribution of the training set, respectively. The worse-than-baseline results imply that our uniform distribution assumption ensures a safe estimation of the biased context effect.

\begin{table}[t]
\centering
\caption{Debiasing comparison results of CCIM~\cite{yang2023context} and the proposed CLEF on the EMOTIC and CAER-S datasets.}
\vspace{-0.2cm}
\resizebox{\linewidth}{!}{%
\begin{tabular}{c|ccc|ccc}
\toprule
\multirow{2}{*}{\textbf{Dataset}} & \multicolumn{3}{c|}{\textbf{EMOT-Net}~\cite{kosti2019context}} & \multicolumn{3}{c}{\textbf{CAER-Net}~\cite{lee2019context}} \\ \cline{2-7} 
                                  & Vanilla   & w/ CCIM  & w/ CLEF         & Vanilla  & w/ CCIM  & w/ CLEF         \\ \midrule
EMOTIC                            & 27.93     & 30.88    & \textbf{31.67}  & 23.85    & 26.51    & \textbf{27.44}  \\
CAER-S                            & 74.51     & 75.82    & \textbf{77.03}  & 73.47    & 74.81    & \textbf{75.86}  \\ \bottomrule
\end{tabular}
}
\label{causal}
\vspace{-6pt}
\end{table}

\noindent \textbf{Debiasing Ability Comparison.}
A gain comparison between our CLEF and the previous CAER debiasing effort CCIM on both datasets is presented in \Cref{causal}. Intuitively, our framework consistently outperforms CCIM~\cite{yang2023context} in both methods. The reasonable reason is that CCIM fails to capture the pure context bias due to over-reliance on the predefined context confounders, causing sub-optimal solutions. In contrast, CLEF decouples the good context prior and the bad context effect, enabling robust debiased predictions.

\subsection{Qualitative Evaluation}
\Cref{vis} shows the performance of vanilla CD-Net before and after counterfactual debiasing via CLEF.
Intuitively, our framework effectively corrects the misjudgments of the vanilla method for emotional states in diverse contexts. Taking \Cref{vis}\textcolor{red}{a} as an example, CLEF eliminates spurious correlations between vegetation-related contexts and positive emotions, giving negative categories aligned with ground truths. Moreover, the CLEF-based CD-Net in \Cref{vis}\textcolor{red}{e} excludes misleading clues about negative emotions provided by dim contexts and achieves an unbiased prediction.

\section{Conclusion}
This paper proposes CLEF, a causal debiasing framework based on counterfactual inference to address the context bias interference in CAER. CLEF reveals that the harmful bias confounds model performance along the direct causal effect via the tailored causal graph, and accomplishes bias mitigation by subtracting the direct context effect from the total causal effect. Extensive experiments prove that CLEF brings favorable improvements to existing models.

\noindent \textbf{Acknowledgements}
This work is supported in part by the National Key R\&D Program of China (2021ZD0113503) and in part by the Shanghai Municipal Science and Technology Committee of Shanghai Outstanding Academic Leaders Plan (No.\,21XD1430300).
 
{
    \small
    \bibliographystyle{ieeenat_fullname}
    \bibliography{main}
}


\end{document}